\documentclass[letterpaper]{article} 
\usepackage{aaai23}  
\usepackage{times}  
\usepackage{helvet}  
\usepackage{courier}  
\usepackage[hyphens]{url}  
\usepackage{graphicx} 
\usepackage{natbib}  
\usepackage{caption} 
\usepackage{algorithm}
\usepackage{algorithmic}
\usepackage{amsmath} 
\usepackage{multirow}
\usepackage{booktabs} 
\usepackage{bm}
\usepackage{amssymb}
\usepackage{stmaryrd}
\usepackage{bbold}
\usepackage{colortbl}
\definecolor{mygray}{gray}{.9}

\newtheorem{definition}{Definition}[section]

\usepackage{newfloat}
\usepackage{listings}
\DeclareCaptionStyle{ruled}{labelfont=normalfont,labelsep=colon,strut=off} 
\floatstyle{ruled}
\newfloat{listing}{tb}{lst}{}
\floatname{listing}{Listing}
\title{Disentangle and Remerge: Interventional Knowledge Distillation for Few-Shot Object Detection from A Conditional Causal Perspective}

\author{
    Jiangmeng Li\textsuperscript{\rm 1}\textsuperscript{\rm 2}\equalcontrib, Yanan Zhang\textsuperscript{\rm 1}\textsuperscript{\rm 2}\equalcontrib, Wenwen Qiang\textsuperscript{\rm 1}\textsuperscript{\rm 2}\thanks{Corresponding author.}, Lingyu Si\textsuperscript{\rm 1}\textsuperscript{\rm 2}, Chengbo Jiao\textsuperscript{\rm 3}, \\Xiaohui Hu\textsuperscript{\rm 2}, Changwen Zheng\textsuperscript{\rm 2}, Fuchun Sun\textsuperscript{\rm 4}
}
\affiliations{
    \textsuperscript{\rm 1}University of Chinese Academy of Sciences\\
    \textsuperscript{\rm 2}Institute of Software Chinese Academy of Sciences\\
    \textsuperscript{\rm 3}University of Electronic Science and Technology of China\\
    \textsuperscript{\rm 4}Tsinghua University\\
    \{jiangmeng2019, yanan2018, qiangwenwen, lingyu, hxh, changwen\}@iscas.ac.cn,\\ chengbojiao@hotmail.com, fcsun@mail.tsinghua.edu.cn
    
}

\usepackage{bibentry}

\begin{document}

\maketitle

\begin{abstract}
Few-shot learning models learn representations with limited human annotations, and such a learning paradigm demonstrates practicability in various tasks, e.g., image classification, object detection, etc. However, few-shot object detection methods suffer from an intrinsic defect that the limited training data makes the model cannot sufficiently explore semantic information. To tackle this, we introduce knowledge distillation to the few-shot object detection learning paradigm. We further run a motivating experiment, which demonstrates that in the process of knowledge distillation, the empirical error of the teacher model degenerates the prediction performance of the few-shot object detection model as the student. To understand the reasons behind this phenomenon, we revisit the learning paradigm of knowledge distillation on the few-shot object detection task from the causal theoretic standpoint, and accordingly, develop a Structural Causal Model. Following the theoretical guidance, we propose a backdoor adjustment-based knowledge distillation method for the few-shot object detection task, namely \textit{\textbf{D}isentangle and \textbf{R}emerge} (D\&R), to perform conditional causal intervention toward the corresponding Structural Causal Model. Empirically, the experiments on benchmarks demonstrate that D\&R can yield significant performance boosts in few-shot object detection. Code is available at \url{https://github.com/ZYN-1101/DandR.git}.
\end{abstract}

\section{Introduction}\label{sec:introduction}

\begin{figure}
    \centering
	{\includegraphics[width=.9\columnwidth]{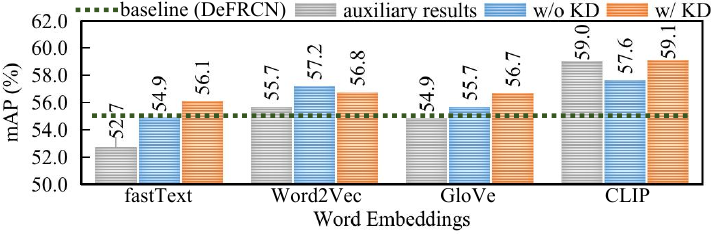}}
	\caption{Comparisons of FSOD models enhanced by auxiliary approaches. Besides the main FSOD task, we introduce an auxiliary task, which encodes the \textit{categories} by the auxiliary approaches and then trains the feature extractor by backpropagating the cross-entropy loss based on the embedded categories and visual features learned by the feature extractor. The inference results are achieved by using schemes: 1) \textit{auxiliary results} are the auxiliary outputs; 2) the results of \textit{w/o KD} are the main outputs; 3) for \textit{w/ KD}, we introduce the knowledge distillation in fine-tuning, and the results are the main outputs. Refer to Figure \ref{fig:frame} for architecture details.}
	\label{fig:motiv}
\end{figure}

Learning robust and generic representations with limited labels is a long-standing topic in machine learning. Few-shot learning, an innovative representation learning paradigm, is practicable in various tasks, e.g., image classification \cite{DBLP:conf/icml/FinnAL17, DBLP:conf/nips/VinyalsBLKW16, DBLP:conf/nips/SnellSZ17, DBLP:conf/cvpr/SungYZXTH18, DBLP:conf/iclr/ChenLKWH19}, object detection \cite{DBLP:conf/iccv/YanCXWLL19, DBLP:conf/iccv/KangLWYFD19, DBLP:conf/icml/WangH0DY20, DBLP:conf/iccv/QiaoZLQWZ21, DBLP:conf/cvpr/ZhuCASS21}, etc.

In general, researchers explore approaches to tackle object detection problems under the setting of few-shot learning in two promising directions: 1) the meta-based methods train the model via a huge amount of few-shot detection tasks sampled from the base classes; 2) the fine-tune-based methods aim to transfer the knowledge from base classes to novel classes. Due to the intrinsic limitation of few-shot learning, it is challenging for the model to sufficiently explore semantic information from the input data. Therefore, we introduce knowledge distillation \cite{DBLP:journals/corr/HintonVD15} to improve the ability of few-shot object detection (FSOD) models to acquire semantic information by learning from large-scale pre-trained models, such as CLIP \cite{DBLP:conf/icml/RadfordKHRGASAM21}. According to the knowledge distillation learning paradigm, minimizing distillation loss entails aligning the distribution of classification \textit{logits} generated by the \textit{teacher} model and \textit{student} model, and thus both the ``correct'' and ``incorrect'' knowledge of the teacher model is learned by the student model in the learned feature space. From the foundational principle of knowledge distillation, the distillation loss can be considered as an auxiliary loss to improve the performance of the main model, and further, in the case that the teacher model has a stronger semantic capturing ability than the student model, the knowledge distillation usually has a considerable promotion on the main model. Yet, observations from the motivating experiments in Figure \ref{fig:motiv}, on the other hand, contradict this.

Specifically, we run the motivating experiments on the VOC dataset \cite{DBLP:journals/ijcv/EveringhamGWWZ10} with Novel Set 1 by using benchmark methods, including fastText \cite{DBLP:conf/emnlp/PenningtonSM14}, Word2Vec \cite{DBLP:conf/nips/MikolovSCCD13}, GloVe \cite{DBLP:conf/emnlp/PenningtonSM14}, and CLIP \cite{DBLP:conf/icml/RadfordKHRGASAM21}. The results are demonstrated in Figure \ref{fig:motiv}. We observe that the variant of w/ KD generally outperforms the compared variants. The auxiliary results, as the control group, are the lowest on most tasks. According to our statement, CLIP, as a large-scale vision-language model, can better improve the model's ability to capture semantic information by using knowledge distillation. However, there exists a counterintuitive phenomenon: for Word2Vec, the w/ KD variant underperforms the w/o KD variant. A plausible explanation is that in the process of knowledge distillation, the FSOD model, as the student model, not only learns the knowledge of the teacher model for the acquisition of open-set semantic information, but also the empirical error of the teacher model degenerates the student model's prediction of the target labels. The teacher's quality severely affects the performance of the student, and several specific teachers may not improve the performance of the student, but instead interfere with the student's predictions on downstream tasks. This is in accordance with the observation of Figure \ref{fig:motiv}. Therefore, such a reason may degenerate the performance of all knowledge distillation-based models, including CLIP-based models. 

To tackle this issue, we revisit the learning paradigm of knowledge distillation on the FSOD task from the causal theoretic standpoint. Accordingly, we develop a Structural Causal Model (SCM) \cite{pearl2009causal, glymour2016causal} \footnote{The principal concepts and methodologies are shared by \cite{pearl2009causal} and \cite{glymour2016causal}.} to describe the causal relationships between the corresponding variables in this paper. As demonstrated in Figure \ref{fig:scm}, the proposed SCM focuses on exploring the causal graph of knowledge distillation-related variables, including the candidate image data, whole open-set semantic knowledge of the teacher model, classification knowledge for downstream tasks, general discriminant knowledge for distinguishing foreground and background objects, and target label. When analyzing the SCM, we discover that it is an exception to current causal inference approaches and that the existing standard definition of the backdoor criterion has limitations, to a certain extent. Inspired by recent works \cite{van2014constructing, perkovic2018complete, correa2017causal}, we propose to expand the backdoor criterion's application boundary on the conditional intervention cases without using extra symbols.

For the detailed methodology, guided by the proposed SCM, we \textit{disentangle} the knowledge distillation objective into four terms. By analyzing the impact of such terms against the SCM, we determine that a specific term can be treated as a \textit{confounder}, which leads the student model to learn the \textit{exceptional} correlation relationship between the classification knowledge and general discriminant knowledge for distinguishing foreground and background objects of the teacher model during knowledge distillation. This is the pivotal reason behind the explanation of the observation in Figure \ref{fig:motiv}, i.e., interfering with the student's predictions. Then, to eliminate the negative impact of the confounder and execute conditional causal intervention toward the proposed SCM, we remove the confounder term and \textit{remerge} the remaining terms as the new knowledge distillation objective. We name the proposed backdoor adjustment-based approach \textit{\textbf{D}isentangle and \textbf{R}emerge} (D\&R). Our experiments on multiple benchmark datasets demonstrate that D\&R can improve the performance of the state-of-the-art FSOD approaches. The sufficient ablation study further proves the effectiveness of the proposed method. Our major contributions are four-fold:
\begin{itemize}
\item We introduce the knowledge distillation to improve the ability of FSOD models to acquire semantic information by learning from large-scale pre-trained models.

\item We observe a paradox that adopting different teacher models, knowledge distillation may both promote and interfere with the prediction of the student model.

\item To understand the causal effects of the knowledge distillation learning paradigm, we establish the SCM. We propose to expand the backdoor criterion's application boundary on the conditional intervention cases without using extra symbols.

\item Guided by the planned SCM, we propose a new method, called Disentangle and Remerge (D\&R), by implementing knowledge distillation with backdoor adjustment. Empirical evaluations demonstrate the superiority of D\&R over state-of-the-art methods.

\end{itemize} 

\section{Related work} \label{sec:relatedwork}

\subsection{Vision-Language Models}
Vision-language models have attracted a lot of attention and shown impressive potential in several areas \cite{DBLP:conf/cvpr/00010BT0GZ18, DBLP:conf/iccv/AntolALMBZP15,DBLP:conf/iccv/HuangWCW19, DBLP:conf/cvpr/YouJWFL16, ma2022open}. High-quality annotated multi-modal data is often difficult to obtain, so unsupervised learning is preferred nowadays. Typical works \cite{DBLP:conf/nips/LuBPL19, DBLP:conf/emnlp/TanB19, DBLP:journals/corr/abs-1909-11740, DBLP:conf/eccv/Li0LZHZWH0WCG20} have made tremendous progress in learning universal representations that 
are easily transferable to downstream tasks via prompting \cite{DBLP:conf/icml/JiaYXCPPLSLD21, DBLP:journals/corr/abs-2010-00747}. CLIP~\cite{DBLP:conf/icml/RadfordKHRGASAM21} is one of the most impressive works, which leverages contrastive learning to align the embedding spaces of texts and images using 400 million image-text pairs, and achieves remarkable performance gain in various tasks. We are the first to introduce CLIP into FSOD.

\subsection{Few-Shot Object Detection}
FSOD aims to build detectors toward limited data scenarios.
Meta-based methods~\cite{DBLP:conf/iccv/YanCXWLL19, DBLP:conf/iccv/KangLWYFD19, DBLP:conf/cvpr/KarlinskySHSAFG19} dominate early research. TFA~\cite{DBLP:conf/icml/WangH0DY20} outperforms the previous meta-based methods by only fine-tuning the last layer of the detector. After that, fine-tune-based methods~\cite{DBLP:conf/eccv/WuL0W20, DBLP:conf/cvpr/ZhangW21a} become popular. The most related works to our approach are SRR-FSD~\cite{DBLP:conf/cvpr/ZhuCASS21} and Morphable Detector (MD)~\cite{DBLP:conf/iccv/ZhaoZW21}, which introduce external information to boost the detection of novel classes. Differently, these two methods adopt pure language models to generate semantic embeddings, bringing bias because of the domain gap. Moreover, our method is able to draw on external information more effectively through the distillation loss we proposed.

\subsection{Knowledge Distillation}
Knowledge distillation is first proposed by the work of ~\cite{DBLP:conf/kdd/BucilaCN06} and ~\cite{DBLP:journals/corr/HintonVD15}. Generally, knowledge distillation can be divided into three categories: logits-based methods~\cite{DBLP:journals/corr/HintonVD15, DBLP:conf/iccv/ChoH19, DBLP:conf/aaai/YangXQY19, zhao2022decoupled}, feature-based methods~\cite{DBLP:journals/corr/RomeroBKCGB14, DBLP:conf/iclr/ZagoruykoK17} and relation-based methods~\cite{DBLP:conf/cvpr/YimJBK17, DBLP:conf/iccv/TungM19}. 
Feature-based methods and relation-based methods achieve preferable performance nowadays. \citet{zhao2022decoupled} which shows competitive results decouples the loss function of the classical logits-based method and provides insights to analyze the key factors of distillation. Guided by the proposed SCM, we disentangle the knowledge distillation objective and remerge them.

\subsection{Causal Inference}
In the past few years, causal inference~\cite{pearl2009causal, glymour2016causal} has been widely applied in various fields such as statistics, economics, and computer science. Specifically, in the area of computer vision, it focuses on eliminating spurious correlations through deconfounding~\cite{DBLP:conf/cvpr/Lopez-PazNCSB17, he2021towards} and counterfactual inference~\cite{DBLP:conf/cvpr/YueWS0Z21, DBLP:conf/cvpr/ChangAG21}. Deconfounding enables estimating causal effects behind confounders. \citet{DBLP:conf/iccv/WangZSZ21} introduces a causal attention module (CaaM) to 
learn causal features with the unsupervised method. CIRL~\cite{DBLP:conf/cvpr/LvLLZLWL22} builds a SCM to formalize the problem of domain generation and separates the causal factors from the non-causal factors in the input data to learn domain-independent representations. We introduce causal inference in FSOD and build the SCM to understand its learning paradigm when applying knowledge distillation. Guided by the SCM, we propose D\&R to boost performance.

\section{Problem Formulation} \label{sec:formulation}

\subsection{Knowledge Distillation for Few-Shot Object Detection} \label{sec:kd}

Under the intuition that the open-set semantic knowledge can support the object detection task in the few-shot setting, we propose to introduce the knowledge distillation approach in the fine-tuning phase of the FSOD model. In particular, the vanilla knowledge distillation \cite{DBLP:journals/corr/HintonVD15} can be formulated as:
\begin{equation}
    \mathcal{L}_\textrm{KD} = \textrm{KL}\left( \mathcal{P}^\mathcal{T} \Vert  \mathcal{P}^\mathcal{S}\right) = \sum_{i=1}^{N^C}{p_i^\mathcal{T}}log\left(\frac{p_i^\mathcal{T}}{p_i^\mathcal{S}}\right),
    \label{eq:kd}
\end{equation}
where $\mathcal{T}$ and $\mathcal{S}$ denote the teacher model and the student model, respectively. ${N^C}$ is the number of categories for the FSOD task (including the ``background'' category). $p_i^\mathcal{T}$ and $p_i^\mathcal{S}$ denote the classification probabilities generated by the corresponding models using the softmax function. $p_i^\mathcal{T}$ and $p_i^\mathcal{S}$, as variables, are sampled \textit{i.i.d} from distributions $\mathcal{P}^\mathcal{T}$ and $\mathcal{P}^\mathcal{S}$, respectively. Note that such a knowledge distillation process is based on the \textit{soft-target} form. We propose to treat the large-scale pre-trained model as the teacher model and the FSOD model as the student model.

\begin{figure}
	\begin{center}
		{\includegraphics[width=1\columnwidth]{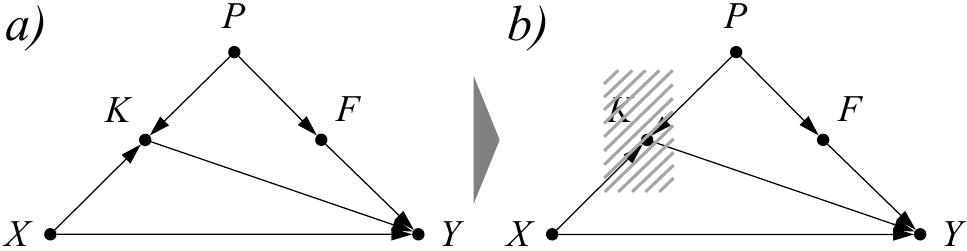}}
		\caption{The proposed SCM between candidate image data $X$, open-set semantic knowledge of the large-scale pre-trained model (e.g., CLIP) $P$, classification knowledge for specific domains $K$, general discriminant knowledge for distinguishing foreground and background objects $F$, and target label $Y$. a) presents the common causal graph, and b) presents the conditional causal graph, where the causal effect is conditional on $K$.}
		\label{fig:scm}
	\end{center}
\end{figure}

\begin{figure*}
	\begin{center}
		{\includegraphics[width=2\columnwidth]{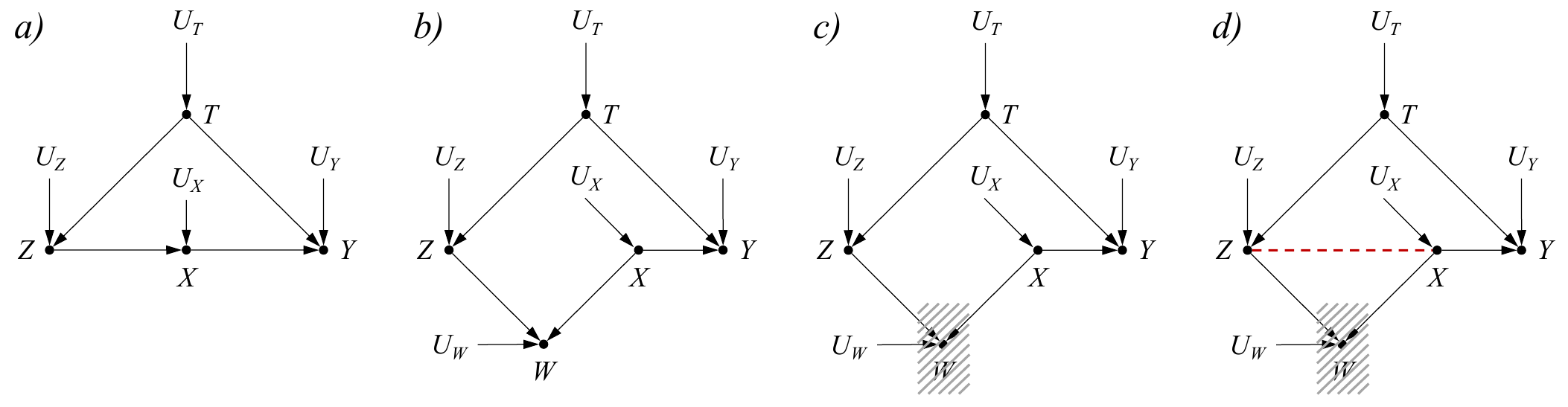}}
		\caption{Examples of the graphical model of SCM. The subfigures a) and b) denote the common cases of performing the intervention on $X$ to explore the causal effects between $X$ and $Y$. The subfigure c) presents a specific case of, given $W$, performing the conditional intervention on $X$ to explore the covariate-specific causal effects between $X$ and $Y$. The red dashed line in the subfigure d) denotes the unordered dependency relationship between $X$ and $Z$. Note that we differentiate the major variables used here and in Figure \ref{fig:scm} to avoid confusion.}
		\label{fig:scmexample}
	\end{center}
\end{figure*}

\subsection{Structural Causal Model} \label{sec:scm}

Minimizing the objective formulated by Equation \ref{eq:kd} is to make the model learn the object detection and classification knowledge from the large-scale pre-trained model in a distillation manner. From this perspective, minimizing the knowledge distillation objective equals aligning $\mathcal{P}^\mathcal{T}$ and $\mathcal{P}^\mathcal{S}$ so that both the ``correct'' and ``incorrect'' knowledge of the teacher model can be learned by the student model. Then, the SCM implicated in the learning paradigm of knowledge distillation is formalized in Figure \ref{fig:scm}. The nodes in SCM represent the abstract information variables, e.g., $X$, and the directed edges represent the (functional) causality, e.g., $X \to Y$ represents that $X$ is the cause and $Y$ is the effect. In the following, we describe the proposed SCM and the rationale behind its construction in detail at a high level.

\bm{$X\to Y \gets K$}. $X$ denotes the candidate image data in a downstream task. $Y$ denotes the corresponding classification label. $K$ denotes the classification knowledge for the specific task. $Y$ is determined by $X$ via two ways: the direct $X \to Y$ and the mediation $X \to K \to Y$. In particular, the first way is the straightforward causal effect. The reasons behind the second way causal effect are: 1) $X \to K$: the domain of a visual dataset is determined by the candidate image data, and thus the corresponding classification knowledge for the domain is determined by the candidate image data. 2) $K \to Y$: the target label can be predicted based on the specific classification knowledge.

\bm{$P \to K \to Y \gets F \gets P$}. We denote $P$ as the open-set semantic knowledge of the large-scale pre-trained model and $F$ as the general discriminant knowledge for distinguishing foreground and background objects. $Y$ is jointly determined by $P$ via the mediation ways, including $P \to K \to Y$ and $P \to F \to Y$. Specifically, 1) $P \to K$: the domain-specific knowledge is extracted from the open-set semantic knowledge. 2) $P \to F$: the knowledge learned by the pre-trained model contains the discriminant knowledge for distinguishing foreground and background objects, because, during pre-training, the input data of the model includes pairs of an image and the corresponding description (or label), and the description focuses on representing features of \textit{foreground} objects so that the pre-trained model contains the general discriminant knowledge for distinguishing foreground and background objects. 3) $F \to Y$: in the target downstream task, i.e., FSOD, the mentioned general discriminant knowledge for distinguishing foreground and background objects is critical to determine the label, since ``\textit{background}'' is a particular label in such an experimental setting. Concretely, for the FSOD task, the open-set semantic knowledge includes the general discriminant knowledge to distinguish foreground and background objects and the domain-specific classification knowledge to specifically classify the foreground objects. Therefore, the mediation causal effect $P \to K \to Y \gets F \gets P$ holds.

The intuition behind our assumption of Figure \ref{fig:scm} b) is that as the domain is fixed for a specific downstream task, the target categories are constant so that the corresponding classification knowledge is determined, and our expected causal effect between $X$ and $Y$ needs to be quantified conditional on $K$. According to the conditional independence theorem of chains in SCM \cite{glymour2016causal}, given $K$, the mediation causal path $X \to K \to Y$ is blocked, i.e., $X$ and $Y$ are \textit{independent} conditional on $K$ in $X \to K \to Y$. However, according to the conditional dependence theorem of colliders in SCM \cite{glymour2016causal}, $X$ and $P$ are \textit{dependent} conditional on $K$ in $X \to K \gets P$ so that if we directly measure the causal effect $X \to Y$, the quantified results may be biased due to $P$. We aim to apply the adjustment approach to quantify the causal effect $X \to Y$ based on the backdoor criterion, yet the common definition of the backdoor path does not apply to the specific SCM case in Figure \ref{fig:scm} b).

\subsection{Discussion on the Backdoor Path} \label{sec:gbp}

\begin{definition}
	\label{def:backdoorcrit}
	(\textbf{The Backdoor Criterion} \cite{glymour2016causal}) Given an ordered pair ($X$, $Y$) in a directed acyclic causal graph $G$, a set of variables $Z$ satisfies the \textbf{backdoor criterion} relative to ($X$, $Y$) if no node in $Z$ is a descendant of $X$, and $Z$ blocks every \textbf{backdoor path} between $X$ and $Y$ having an arrow into $X$.
\end{definition}

According to Definition \ref{def:backdoorcrit}, \cite{glymour2016causal} proposes the common definition of the backdoor path to demarcate the scope of application of the backdoor criterion. Such a backdoor path definition can be applied in most cases, e.g., the common cases demonstrated in Figure \ref{fig:scmexample} a) and b). However, the common backdoor path definition cannot be applied in the case of conditional intervention. For instance, as shown in Figure \ref{fig:scmexample} c) and d), given $W$, we aim to explore the covariate-specific causal effects between $X$ and $Y$. As the conditional dependence theorem of colliders in SCM, if a collider node, i.e., one node receiving edges from two other nodes, exists, conditioning on the collision node produces an unordered dependence between the node's parents. Therefore, the causal path $Y \gets T \to Z \to W \gets X$ originally blocked by the collider node, in Figure \ref{fig:scmexample} b), is connected conditional on $W$, in Figure \ref{fig:scmexample} c). According to the common backdoor path definition in Definition \ref{def:backdoorcrit}, the connected path $Y \gets T \to Z \to W \gets X$ is still not a backdoor path, but the confounder $T$ impacts both $X$ and $Y$ so that the true covariate-specific causal effects between $X$ and $Y$ cannot be directly calculated.

To tackle this issue, recent works \cite{van2014constructing, perkovic2018complete, correa2017causal} are committed to exploring \textit{updated} SCMs to determine how to impose the backdoor adjustment in different scenarios, yet they require building a new SCM by using more complex symbology and case-specific analyses. Inspired by such approaches, we propose to expand the backdoor criterion's application boundary on the conditional intervention cases without using extra symbols, which shares the intrinsic intuition with \cite{van2014constructing, perkovic2018complete, correa2017causal}. In detail, given an ordered pair of variables ($X$, $Y$) in a directed acyclic structural causal graph $G$, a path satisfies the definition of the backdoor path relative to ($X$, $Y$) if it contains a \textit{confounder} $T$ that jointly infers both $X$ and $Y$, e.g., for $T$ and $X$, $T$ is the cause of $X$, or $T$ and $X$ are dependent if no direct causal relationship exists.

As shown in Figure \ref{fig:scmexample} d), $T$ has direct ordered causal relationships with $Z$ and $Y$. $Z$ and $X$ are dependent without a direct causal relationship as denoted by the red dashed line in Figure \ref{fig:scmexample} d). Therefore, $T$ is the shared cause of $X$ and $Y$. $T$ can be treated as a \textit{confounder}, and the path $Y \gets T \to Z \to W \gets X$ is a backdoor path conditional on $W$. We can achieve the true covariate-specific causal effects between $X$ and $Y$ by performing the backdoor adjustment.

\begin{figure*}
	\begin{center}
		{\includegraphics[width=1.9\columnwidth]{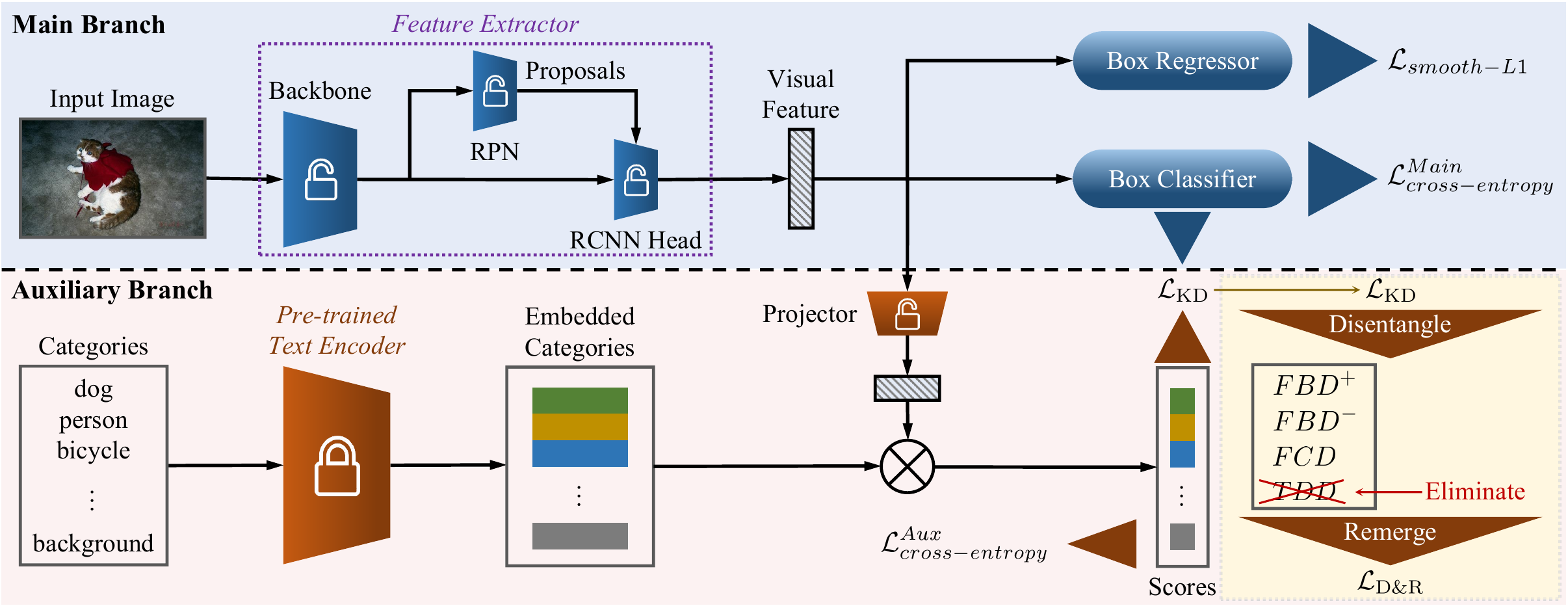}}
		\caption{D\&R's architecture. We introduce an auxiliary branch to the main branch of the FSOD benchmark approach. The $d$-dimension embedded categories are obtained from a frozen pre-trained text encoder. A linear projector is applied to transform the visual feature into $d$-dimension vectors. While training base classes, we backpropagate $\mathcal{L}^{Main}_{cross-entropy}$, $\mathcal{L}^{Aux}_{cross-entropy}$ and $\mathcal{L}_{smooth-L1}$ to train the feature extractor and projector in an end-to-end manner. We introduce $\mathcal{L}_{\textrm{D\&R}}$ to promote the detection during fine-tuning. 
		}
		\label{fig:frame}
	\end{center}
\end{figure*}

\subsection{Conditional Causal Intervention via Backdoor Adjustment} \label{sec:cci}

An ideal FSOD model should capture the true causality between $X$ and $Y$ and can generalize to unseen samples well. For the knowledge distillation empowered training approach, as shown in Figure \ref{fig:scm}, we expect to capture the direct causal relationship between $X$ and $Y$ independent of $P$. However, from the proposed SCM demonstrated in Figure \ref{fig:scm} b), the increased likelihood of $Y$ given $X$ is not only due to $X \to Y$, but also the spurious correlation via $X \to K \gets P \to F \to Y$ conditional on $K$. Consequently, the prediction of $Y$ is based on not only the input data $X$, but also the semantic knowledge taught by the pre-trained model, which is demonstrated by the experiments in Figure \ref{fig:motiv}. Therefore, to pursue the true causality between $X$ and $Y$, we need to use the conditional causal intervention $P( {Y(X)| {do(X)} } )$ instead of the $P( {Y(X)| {X} } )$.

We propose to use the backdoor adjustment \cite{glymour2016causal} to eliminate the interference of different teachers' knowledge. The backdoor adjustment assumes that we can observe and adjust the set of variables satisfying the backdoor criterion to achieve the true causal effect with intervention. In the proposed SCM, the semantic knowledge contained in $P$ is immeasurable, because the input data domain is constant for a specific task. However, the general discriminant knowledge for distinguishing foreground and background objects contained in $F$ is shared among different tasks so that we can observe and adjust $F$ to achieve the true causality between $X$ and $Y$. Formally, the backdoor adjustment for the proposed SCM is presented as:
\begin{equation}
\label{eq:ba}
P( {Y( X )| {do( X )} } ) = \sum\limits_{j = 1}^{N^F} {P( {Y( {X} )| {X, \hat{F}_j} } )P( {\hat{F}_j} )},
\end{equation}
where $P( {Y( {X} )| {do( {X} )} } )$ represents the true causality between $X$ and $Y$, and $\hat{F}_j$ denotes the stratified knowledge of $F$, i.e., $F = \left\{\hat{F}_j \big| j \in \llbracket {1, N^F} \rrbracket \right\}$. 

\begin{table*}[t]
	\setlength{\tabcolsep}{3.9pt}
	\begin{center}
		\begin{small}
			\begin{tabular}{l|ccccc|ccccc|ccccc}
        		\toprule
        		\multirow{2}*{Methods} &  
        		\multicolumn{5}{c|}{Novel Set 1} &
        		\multicolumn{5}{c|}{Novel Set 2} &
        		\multicolumn{5}{c}{Novel Set 3}  \\
                & 1 & 2 & 3 & 5 & 10 
                & 1 & 2 & 3 & 5 & 10 
                & 1 & 2 & 3 & 5 & 10 \\
                \midrule
        	\multicolumn{16}{c}{\textbf{Results of single run, following TFA~\cite{DBLP:conf/icml/WangH0DY20}}} \\ \midrule
	FSRW~\cite{DBLP:conf/iccv/KangLWYFD19} & 14.8 & 15.5 & 26.7 & 33.9 & 47.2 & 15.7 & 15.3 & 22.7 & 30.1 & 40.5 & 21.3 & 25.6 & 28.4 & 42.8 & 45.9 \\
        		TFA~\cite{DBLP:conf/icml/WangH0DY20} & 39.8 & 36.1 & 44.7 & 55.7 & 56.0 & 23.5 & 26.9 & 34.1 & 35.1 & 39.1 & 30.8 & 34.8 & 42.8 & 49.5 & 49.8 \\
        		MPSR~\cite{DBLP:conf/eccv/WuL0W20} & 41.7 & 42.5 & 51.4 & 55.2 & 61.8 & 24.4 & 29.3 & 39.2 & 39.9 & 47.8 & 35.6 & 41.8 & 42.3 & 48.0 & 49.7 \\
        		FSCE~\cite{DBLP:conf/cvpr/SunLCYZ21} & 44.2 & 43.8 & 51.4 & 61.9 & 63.4 & 27.3 & 29.5 & 43.5 & 44.2 & 50.2 & 37.2 & 41.9 & 47.5 & 54.6 & 58.5 \\
        		SRR-FSD\ddag~\cite{DBLP:conf/cvpr/ZhuCASS21} & 47.8 & 50.5 & 51.3 & 55.2 & 56.8 & 32.5 & 35.3 & 39.1 & 40.8 & 43.8 & 40.1 & 41.5 & 44.3 & 46.9 & 46.4 \\
        		
        		Meta Faster R-CNN~\cite{DBLP:conf/aaai/HanHMHC22} & 43.0 & 54.5 & 60.6 & \textbf{66.1} & 65.4 & 27.7 & 35.5 & 46.1 & 47.8 & 51.4 & 40.6 & 46.4 & 53.4 & 59.9 & 58.6 \\
        		FCT~\cite{Han_2022_CVPR} & 49.9 & 57.1 & 57.9 & 63.2 & \textbf{67.1} & 27.6 & 34.5 & 43.7 & 49.2 & 51.2 & 39.5 & 54.7 & 52.3 & 57.0 & 58.7 \\
        		\citet{Kaul_2022_CVPR} & 54.5 & 53.2 & 58.8 & 63.2 & 65.7 & 32.8 & 29.2 & \textbf{50.7} & 49.8 & 50.6 & 48.4 & 52.7 & 55.0 & 59.6 & 59.6 \\
        		\hline	
        		DeFRCN*~\cite{DBLP:conf/iccv/QiaoZLQWZ21} & 55.1 & 61.9 & 64.9 & 65.8 & 66.2 & 33.8 & 45.1 & 46.1 & \textbf{53.2} & 52.3 & 51.0 & 56.6 & 55.6 & 59.7 & \textbf{61.9} \\
        		D\&R (Ours)\ddag & \textbf{60.4} & \textbf{64.0} & \textbf{65.2} & 64.7 & 66.3 & \textbf{37.9} & \textbf{46.8} & 48.1 & 52.7 & \textbf{53.1} & \textbf{55.7} & \textbf{57.9} & \textbf{57.6} & \textbf{60.6} & \textbf{61.9} \\ 
        		\midrule
        		\multicolumn{16}{c}{\textbf{Average results of 30 runs, following TFA~\cite{DBLP:conf/icml/WangH0DY20}}} \\ \midrule
        		FRCN+ft-full~\cite{DBLP:conf/iccv/YanCXWLL19} & 9.9 & 15.6 & 21.6 & 28.0 & 35.6 & 9.4 & 13.8 & 17.4 & 21.9 & 29.8 & 8.1 & 13.9 & 19.0 & 23.9 & 31.0 \\
                Xiao et al~\cite{DBLP:conf/eccv/XiaoM20} & 24.2 & 35.3 & 42.2 & 49.1 & 57.4 & 21.6 & 24.6 & 31.9 & 37.0 & 45.7 & 21.2 & 30.0 & 37.2 & 43.8 & 49.6 \\
        		TFA~\cite{DBLP:conf/icml/WangH0DY20} & 25.3 & 36.4 & 42.1 & 47.9 & 52.8 & 18.3 & 27.5 & 30.9 & 34.1 & 39.5 & 17.9 & 27.2 & 34.3 & 40.8 & 45.6 \\
        		FSCE~\cite{DBLP:conf/cvpr/SunLCYZ21}& 32.9 & 44.0 & 46.8 & 52.9 & 59.7 & 23.7 & 30.6 & 38.4 & 43.0 & 48.5 & 22.6 & 33.4 & 39.5 & 47.3 & 54.0 \\
        		DCNet~\cite{DBLP:conf/cvpr/HuBLCW21} & 33.9 & 37.4 & 43.7 & 51.1 & 59.6 & 23.2 & 24.8 & 30.6 & 36.7 & 46.6 & 32.3 & 34.9 & 39.7 & 42.6 & 50.7 \\
        		FCT~\cite{Han_2022_CVPR} & 38.5 & 49.6 & 53.5 & 59.8 & 64.3 & 25.9 & 34.2 & 40.1 & 44.9 & 47.4 & 34.7 & 43.9 & 49.3 & 53.1 & 56.3 \\
        		\hline
        		DeFRCN*~\cite{DBLP:conf/iccv/QiaoZLQWZ21}& 39.3 & 50.9 & 55.3 & \textbf{61.8} & 65.3 & 	27.4 & 36.8 & 40.4 & 45.1 & 50.8 & 	35.0 & 45.1 & 50.2 & 55.7 & 58.9 \\
        		D\&R (Ours)\ddag & \textbf{41.0} & \textbf{51.7} & \textbf{55.7} & \textbf{61.8} & \textbf{65.4} & 	\textbf{30.7} & \textbf{39.0} & \textbf{42.5} & \textbf{46.6} & \textbf{51.7} & \textbf{37.9} & \textbf{47.1} & \textbf{51.7} & \textbf{56.8} & \textbf{59.5} \\
        		\bottomrule
        	\end{tabular}
		\end{small}
	\end{center}
	\caption{FSOD results (\%) on VOC. * denotes the method re-implemented with one single GPU. \ddag indicates the methods using external knowledge.}
	\label{tab:voc}
\end{table*}
\section{Methodology} \label{sec:method}
\subsection{Overview}

We provide the functional implementations in Figure \ref{fig:frame}. To illustrate our method more clearly, we present the training and fine-tuning paradigm of D\&R in Algorithm \ref{alg:pipeline}. We adopt the two-stage training scheme following DeFRCN~\cite{DBLP:conf/iccv/QiaoZLQWZ21}. In the first stage, we train the detector with abundant samples of base classes. Besides the loss terms in DeFRCN, i.e., $\mathcal{L}_{\textrm{RPN}}$ and $\mathcal{L}_{\textrm{RCNN}}$, we introduce the conventional cross-entropy loss $\mathcal{L}_{cross-entropy}^{Aux}$ to guide the training of the feature extractor and the projector (refer to Figure \ref{fig:frame}). While fine-tuning the network with samples of all categories, i.e., base categories and novel categories, under the generalized few-shot object detection setting (G-FSOD), we propose $\mathcal{L}_{\textrm{D\&R}}$ to boost the performance of the main detection branch. We elaborate on details of $\mathcal{L}_{\textrm{D\&R}}$ in Sections~\ref{sec:kdba} and \ref{sec:dnr}. After training and fine-tuning, the teacher is abandoned, and the main branch is used to produce detection results. 
\begin{algorithm}[tb]
\renewcommand{\algorithmicrequire}{\textbf{Input:}}
\renewcommand\algorithmicensure {\textbf{Output:} }
\caption{D\&R Training and Fine-tuning Paradigm}
\label{alg:pipeline}

\begin{algorithmic}[1]

\REQUIRE ~~\\ 
\text{\#}: $N$, ~minibatch size \\
\text{\#}: $f_{base}$, $f_{all}$, ~detectors \\
\text{\#}: $\lambda$, ~hyper-parameter, the weight of $\mathcal{L}_\textrm{D\&R}$ \\
\text{\#}: $lr_{base}$, $lr_{all}$, ~learning rates \\

\STATE $\# \ training \ on \ samples \ of \ base \ classes$
\REPEAT
\STATE Iteratively sample minibatch ${X_{base}} = \left\{ {{X_i}} \right\}_{i = 1}^N$.
\STATE $\mathcal{L}_{base} \leftarrow \mathcal{L}_{\textrm{RPN}}+\mathcal{L}_{\textrm{RCNN}}+\mathcal{L}^{Aux}_{cross-entropy}$
\STATE $f_{base} \leftarrow f_{base} - lr_{base}{\nabla _f}{\mathcal{L}_{base}}$ \\
\UNTIL $f_{base}$ converge.
\STATE $\# \ fine$-$tuning \ on \ samples \ of \ all \ classes$
\STATE Initialize $f_{all}$ with the weight of converged $f_{base}$.
\REPEAT
\STATE Iteratively sample minibatch ${X_{all}} = \left\{ {{X_i}} \right\}_{i = 1}^N$.
\STATE $\mathcal{L}_{all} \leftarrow \mathcal{L}_{\textrm{RPN}}+\mathcal{L}_{\textrm{RCNN}}+\mathcal{L}^{Aux}_{cross-entropy}+\lambda \mathcal{L}_{\textrm{D\&R}}$
\STATE $f_{all} \leftarrow f_{all} - lr_{all}{\nabla _f}{\mathcal{L}_{all}}$
\UNTIL $f_{all}$ converge.
\end{algorithmic}
\end{algorithm}
\subsection{Knowledge Distillation with Backdoor Adjustment} \label{sec:kdba}

We present the implementation of the backdoor adjustment during the fine-tuning phase. As shown in Equation \ref{eq:ba}, we provide the detailed functional implementations for the knowledge distillation with backdoor adjustment as follows. The foundational idea behind the knowledge distillation is aligning the classification probabilities generated by the teacher and student models (the teacher model is fixed while the student model is trainable) in order to promote the student model to learn both ``correct'' and ``incorrect'' knowledge from the teacher model. Therefore, we represent the functional implementations of the ${P( {Y( {X} )| {X, \hat{F}_j} } )}$ by adopting the loss defined in Equation \ref{eq:kd}:
\begin{equation}
\begin{aligned}
&P( {Y( {X} )| {X, \hat{F}_j} } ) = \\&\sum_{i=1}^{N^C}\left(\lfloor p_i^\mathcal{T}\rceil^{\bar{K}} log\frac{\lfloor p_i^\mathcal{T}\rceil^{\bar{K}}}{\lfloor p_i^\mathcal{S}\rceil^{\bar{K}}} + \lfloor p_i^\mathcal{T}\rceil^{\hat{F}_j} log\frac{\lfloor p_i^\mathcal{T}\rceil^{\hat{F}_j}}{\lfloor p_i^\mathcal{S}\rceil^{\hat{F}_j}}\right),
\end{aligned}
\label{eq:kdba1}
\end{equation}
where $\bar{K}$ denotes the restricted classification knowledge extracted from $K$ due to the data domain of a specific task. $\lfloor p_i^\mathcal{T}\rceil^{\bar{K}}$ and $\lfloor p_i^\mathcal{S}\rceil^{\bar{K}}$ denote the distillation of the knowledge related to $\bar{K}$, and $\lfloor p_i^\mathcal{T}\rceil^{\hat{F}_j}$ and $\lfloor p_i^\mathcal{S}\rceil^{\hat{F}_j}$ denote the distillation of the stratified knowledge related to $F$ in SCM. As a result, we implement the overall backdoor adjustment by
\begin{equation}
\begin{aligned}
&P( {Y( {X} )| {do( {X} )} } ) = \\&\sum_{i=1}^{N^C}\sum_{j=1}^{N^F}\left(\lfloor p_i^\mathcal{T}\rceil^{\bar{K}} log\frac{\lfloor p_i^\mathcal{T}\rceil^{\bar{K}}}{\lfloor p_i^\mathcal{S}\rceil^{\bar{K}}} + \lfloor p_i^\mathcal{T}\rceil^{\hat{F}_j} log\frac{\lfloor p_i^\mathcal{T}\rceil^{\hat{F}_j}}{\lfloor p_i^\mathcal{S}\rceil^{\hat{F}_j}}\right).
\end{aligned}
\label{eq:kdba2}
\end{equation}

According to the proposed SCM, the restricted classification knowledge $\bar{K}$ is related to the specific FSOD task so that $\bar{K}$ is naturally contained in the adjusted knowledge distillation objective, i.e., Equation \ref{eq:kdba2}. Following the principle of backdoor adjustment, we further sum up all conditional causal effects based on adjustment for the stratified knowledge $\hat{F}_j$ in Equation \ref{eq:kdba2}. Therefore, we \textit{disentangle} the knowledge distillation objective and eliminate the \textit{confounder} term and then \textit{remerge} the remaining terms according to the proposed backdoor adjustment methodology.

\subsection{Disentangle and Remerge} \label{sec:dnr}

Inspired by \cite{zhao2022decoupled}, we disentangle the knowledge distillation objective for FSOD into four terms: 1) two terms for positive samples, i.e., the labels of target samples belong to the foreground categories, including \textit{positive Foreground and Background knowledge Distillation} (FBD$^+$) and \textit{Target Discrimination knowledge Distillation} (TDD); 2) one term for negative samples, where the shared label of target samples is ``background'', i.e., \textit{negative Foreground and Background knowledge Distillation} (FBD$^-$); 3) a common term for all samples, i.e., \textit{Foreground Classification knowledge Distillation} (FCD).

FBD$^+$ and FBD$^-$ present the distillation objective of stratified knowledge $\hat{F}$ extracted from general discriminant knowledge for distinguishing foreground and background objects $F$, respectively. Specifically, FBD$^+$ is the disentangled knowledge distillation objective to measure the similarity between the teacher's and student's binary probabilities of the ``background'' category and \textit{non-target} foreground category group. FBD$^-$ is the objective to measure the similarity between the teacher's and student's binary probabilities of the \textit{target} ``background'' category and foreground category group. TDD is the considered confounder, which denotes the objective to measure the similarity between the teacher's and student's binary probabilities of the \textit{target} category and the \textit{non-target} category group. Our explanation of the observation of Figure \ref{fig:motiv} states that in the process of knowledge distillation, the empirical error of the teacher model degenerates the student model's prediction of the target labels, since there exists a confounder leading the student model to learn the exceptional correlation relationship between the classification knowledge and general discriminant knowledge for distinguishing foreground and background objects of the teacher model during knowledge distillation (see Section \ref{sec:introduction} for details). FCD represents the distillation objective of the restricted classification knowledge $\bar{K}$, which is the objective to measure the similarity between the teacher's and student's multiple probabilities among \textit{non-target} foreground categories.

To perform the expected knowledge distillation with backdoor adjustment, we eliminate the confounder TDD and further remerge the remaining disentangled objective terms. According to Equation \ref{eq:kd} and Equation \ref{eq:kdba2}, we derive the final loss function for the proposed D\&R:
\begin{equation}
    \label{eq:ldr}
    \begin{aligned}
    \mathcal{L}_\textrm{D\&R} = &\alpha \textrm{KL}\left( \mathcal{P}^\mathcal{T}_{\textrm{FBD}^+} \Vert  \mathcal{P}^\mathcal{S}_{\textrm{FBD}^+}\right) \\&+ \beta \textrm{KL}\left( \mathcal{P}^\mathcal{T}_{\textrm{FBD}^-} \Vert  \mathcal{P}^\mathcal{S}_{\textrm{FBD}^-}\right) + \textrm{KL}\left( \mathcal{P}^\mathcal{T}_{\textrm{FCD}} \Vert  \mathcal{P}^\mathcal{S}_{\textrm{FCD}}\right) \\= &\sum_{i=1}^{N^C}\Bigg(\alpha \lfloor p_i^\mathcal{T}\rceil^{\hat{F}_1} log\frac{\lfloor p_i^\mathcal{T}\rceil^{\hat{F}_1}}{\lfloor p_i^\mathcal{S}\rceil^{\hat{F}_1}} + \beta \lfloor p_i^\mathcal{T}\rceil^{\hat{F}_2} log\frac{\lfloor p_i^\mathcal{T}\rceil^{\hat{F}_2}}{\lfloor p_i^\mathcal{S}\rceil^{\hat{F}_2}} \\&+ \lfloor p_i^\mathcal{T}\rceil^{\bar{K}} log\frac{\lfloor p_i^\mathcal{T}\rceil^{\bar{K}}}{\lfloor p_i^\mathcal{S}\rceil^{\bar{K}}} \Bigg),
    \end{aligned}
\end{equation}
where $\hat{F}_1$ and $\hat{F}_2$ denote the stratified knowledge of $F$, corresponding to FBD$^+$ and FBD$^-$, respectively. $\alpha$ and $\beta$ are coefficients that control the impact of the terms for positive samples and negative samples in knowledge distillation. 

\section{Experiments}\label{sec:experiments}
\subsection{Experimental Setting} \label{expsetting}
\subsubsection{Benchmarks.}We benchmark D\&R on Pascal VOC~\cite{DBLP:journals/ijcv/EveringhamGWWZ10} and COCO~\cite{DBLP:conf/eccv/LinMBHPRDZ14} datasets following the previous experimental settings~\cite{DBLP:conf/icml/WangH0DY20, DBLP:conf/iccv/QiaoZLQWZ21} for a fair comparison. For Pascal VOC, 15 classes are randomly selected as base classes, and the remaining 5 classes are novel classes. Each novel class has $K = 1, 2, 3, 5, 10$ annotated training samples. We train the network with VOC07 and VOC12 train/val set, and evaluate our method with VOC07 test set using AP$_{50}$ as the evaluation metric. For COCO, there are 60 base categories that are disjoint with VOC and 20 novel classes. Each novel class has $K = 1, 2, 3, 5, 10, 30$ samples. We report COCO-style mAP of novel classes for COCO.
\subsubsection{Implementation Details.} Our model is built upon the state-of-the-art method DeFRCN~\cite{DBLP:conf/iccv/QiaoZLQWZ21} with a backbone network ResNet-101. We use SGD as the optimizer with a batch size of 8. All models are trained with a single GPU. Due to the change in batch size, the number of training iterations is doubled based on the implementation of DeFRCN, and the learning rate is halved. Other parameters are exactly the same as DeFRCN. We add four additional hyper-parameters. 
For the experiments of COCO, the distillation temperature is 5, and the weight of the distillation loss is 5. $\alpha$ and $\beta$ in $\mathcal{L}_\textrm{D\&R}$ are 4 and 0.5, respectively. For the experiments on Pascal VOC, we set the temperature to 10 and the distillation loss weight to 1 empirically. $\alpha$ and $\beta$ are 10 and 2. Moreover, the loss terms of positive and negative samples are averaged separately in the calculation of $\mathcal{L}_\textrm{D\&R}$ to achieve a balance. Our method is implemented on Pytorch 1.9. Results on the COCO dataset are obtained using a single Tesla V100 GPU with a memory of 32G. The experiments on the VOC dataset are conducted using one Geforce RTX3090 GPU with a memory of 24G. The operating system we use is Ubuntu18.04.
\subsection{Comparison Results}
We report $AP_{50}$ for novel classes on three data splits of Pascal VOC in Table \ref{tab:voc}, and the results of COCO-style $mAP$ are shown in Table \ref{tab:coco} for COCO. We observe that D\&R achieves state-of-the-art performance on most tasks. Especially, D\&R has more impressive improvements with fewer annotated samples. At higher shots, D\&R can also obtain competitive results. Specifically, D\&R is, on average, 1.58\% higher than the best baseline method on the VOC dataset. D\&R averagely beats the best benchmark method by 0.68\% on the COCO dataset. COCO is a larger dataset, and Pascal VOC is smaller with respect to the category number and the sample size. Benchmark approaches and D\&R achieve relatively consistent performance on COCO, and as shown in Table \ref{tab:coco}, D\&R's improvements are more consistent. 

Furthermore, we report the average results of multiple repeated runs over different training samples. For Pascal VOC, as the few-shot detection performance on VOC is quite unstable, we set a fixed random seed for all experiments to stabilize the results. Moreover, results reported in DeFRCN~\cite{DBLP:conf/iccv/QiaoZLQWZ21} are produced with multi-GPUs. As the results are extremely different when a single GPU is used, we re-produce the results of DeFRCN~\cite{DBLP:conf/iccv/QiaoZLQWZ21} with one GPU based on the officially released base model and mark the results in Table \ref{tab:voc} with *. From Table \ref{tab:voc}, we learn that our proposed D\&R improves the performance by 2.6\%, 1.7\%, 1.3\%, 0.9\%, and 0.5\% on average when the shot number $K=1,2,3,5,10$, respectively. For COCO, we report the average results of 10 repeated runs of different training samples in Table \ref{tab:coco}. At lower shots, our D\&R has consistent improvements compared with the baseline method DeFRCN~\cite{DBLP:conf/iccv/QiaoZLQWZ21}. When the shot number is 10 or higher, knowledge from CLIP cannot provide much help. This is in agreement with the results on VOC.

\begin{table}[t]
	\setlength{\tabcolsep}{3.pt}
	\begin{center}
		\begin{small}
			\begin{tabular}{l|cccccc}
        		\toprule  
        		\multirow{2}*{Methods} & \multicolumn{6}{c}{Shot Number} \\
        		& 1 & 2 & 3 & 5 & 10 & 30 \\
                \midrule
               \multicolumn{7}{c}{\textbf{Results of single run, following TFA~\cite{DBLP:conf/icml/WangH0DY20}}} \\ \midrule FSRW~\cite{DBLP:conf/iccv/KangLWYFD19} & - & - & - & - & 5.6 & 9.1 \\
                TFA~\cite{DBLP:conf/icml/WangH0DY20} & 3.4 & 4.6 & 6.6 & 8.3 & 10.0 & 13.7 \\
                MPSR~\cite{DBLP:conf/eccv/WuL0W20} & 2.3 & 3.5 & 5.2 & 6.7 & 9.8 & 14.1 \\ 
                FSCE~\cite{DBLP:conf/cvpr/SunLCYZ21} & - & - & - & - & 11.9 & 16.4 \\
                SRR-FSD\ddag~\cite{DBLP:conf/cvpr/ZhuCASS21} & - & - & - & - & 11.3 & 14.7 \\
        		
        		FCT~\cite{Han_2022_CVPR} & 5.6 & 7.9 & 11.1 & 14.0 & 17.1 & 21.4 \\
        		\hline
        		DeFRCN~\cite{DBLP:conf/iccv/QiaoZLQWZ21} & 6.5 & 11.8 & 13.4 & 15.3 & 18.6 & \textbf{22.5} \\
        		D\&R (Ours)\ddag & \textbf{8.3} & \textbf{12.7} & \textbf{14.3} & \textbf{16.4} & \textbf{18.7} & 21.8 \\
        		\midrule
        		\multicolumn{7}{c}{\textbf{Average results of 10 runs, following TFA~\cite{DBLP:conf/icml/WangH0DY20}}} \\ \midrule
        		FRCN+ft-full & 1.7 & 3.1 & 3.7 & 4.6 & 5.5 & 7.4 \\
        		TFA & 1.9 & 3.9 & 5.1 & 7.0 & 9.1 & 12.1 \\
        		\hline
        		DeFRCN & 4.8 & 8.5 & 10.7 & 13.5 & \textbf{16.7} & \textbf{21.0} \\
        		D\&R (Ours)\ddag & \textbf{6.1} & \textbf{9.5} & \textbf{11.5} & \textbf{13.9} & 16.4 & 20.0 \\
        		\bottomrule
        	\end{tabular}
		\end{small}
	\end{center}
	\caption{FSOD results (\%) on COCO. \ddag indicates the methods using external knowledge.}
	\label{tab:coco}
\end{table}

\begin{table}[t]
	\setlength{\tabcolsep}{1.pt}
	\begin{center}
		\begin{small}
			\begin{tabular}{ccccc|cccccc}
        		\toprule  
        		\multicolumn{5}{c|}{Components} &
        		\multicolumn{6}{c}{Shot Number} \\
                KD & TDD & FBD$^+$ & FBD$^-$ & FCD & 1 
                & 2 & 3 & 5 & 10 & 30 \\
                \midrule
                & & & & & 6.77 & 10.94 & 12.96 & 15.22 & 18.02 & 21.63 \\
        		\checkmark & & & & & 7.69 & 11.96 & 13.73 & 15.82 & 18.59 & \textbf{21.84} \\
        		 & & \checkmark & \checkmark & & 7.71 & 11.89 & 13.62 & 15.53 & 18.11 & 21.41 \\
        		 & & & & \checkmark & 8.10 & 12.33 & 14.04 & 16.20 & \textbf{18.70} & 21.76 \\
        		 & \checkmark & \checkmark & \checkmark & \checkmark & 8.05 & 12.40 & 14.26 & 16.34 & \textbf{18.70} & 21.67 \\
        		 & & & \checkmark & \checkmark & 7.91 & 12.22 & 13.94 & 16.07 & 18.59 & 21.72 \\
        		 & & \checkmark & & \checkmark & 8.25 & 12.67 & 14.10 & \textbf{16.50} & 18.55 & 21.81 \\
        		 \rowcolor{mygray}
        		 & & \checkmark & \checkmark & \checkmark & \textbf{8.29} & \textbf{12.71} & \textbf{14.27} & 16.43 & 18.65 & \underline{21.82} \\
        		\bottomrule
        	\end{tabular}
		\end{small}
	\end{center}
	\caption{FSOD results (\%) of the ablation experiments on COCO. The first row indicates the model without the knowledge distillation, and the second row indicates the model with the vanilla knowledge distillation.}
	\label{tab:table_ablation}
\end{table}

\subsection{Ablation Study}
\subsubsection{Effectiveness of $\mathcal{L}_{\textrm{D\&R}}$.}We conduct the ablation experiments to take a closer look at D\&R in Table \ref{tab:table_ablation}. We find that D\&R, shown in the last row, outperforms all ablation variants, and the performance gains mainly owe to the reliable knowledge provided by $\mathcal{L}_{\textrm{D\&R}}$, which proves the effectiveness of D\&R. Although the FBD$^-$ + FCD variant even underperforms the sole FCD variant, according to the backdoor adjustment-based approach, FBD$^-$ can improve the performance of our model (as shown in the last row). Such observations demonstrate the effectiveness of the proposed backdoor adjustment-based learning paradigm. In most cases, combining FBD$^+$, FBD$^-$, and FCD achieves preferable results, but adding TDD may degenerate the performances, e.g., comparing the fifth row and the last row, we observe that TDD indeed degenerates the model's performance, which proves our statement treating TDD as an unexpected confounder, thereby demonstrating the empirical effectiveness of D\&R. Regarding the results, we have a further observation: as the shot number grows, the improvement brought by the distillation loss becomes limited, including $\mathcal{L}_{\textrm{D\&R}}$. The reason is that the semantic information of novel classes is extremely scarce at lower shots, so the knowledge from pre-trained models can effectively improve models. However, when more training samples are available, the knowledge distillation provides less additional information, and the performance improvement is limited.

\subsubsection{Analysis in Training Consumption.}On COCO, during base training, DeFRCN needs 0.834s/iter with a memory of 10159M, while D\&R costs 0.839s/iter with 10179M. For fine-tuning, DeFRCN costs 0.749s/iter with 10008M, and D\&R costs 0.794s/iter with 10040M. The inference time of both methods is 0.08s/image. The training iterations are the same. The results on VOC are consistent. Overall, our approach adds negligible time and space consumption during training and brings no extra consumption during testing.

\subsubsection{Hyper-Parameters}\label{app:hyper-parameters}
There are four important hyper-parameters in $\mathcal{L}_{\textrm{D\&R}}$: distillation temperature $\textrm{T}$, distillation loss weight $\lambda$, and two parameters $\alpha$ and $\beta$ for the weights of FBD$^+$ and FBD$^-$ in $\mathcal{L}_{\textrm{D\&R}}$, respectively. We analyze the effectiveness of different hyper-parameters one by one.

\underline{Distillation Temperature.} We distill the FSOD model using different temperatures with vanilla knowledge distillation on the COCO dataset. As shown in Table \ref{tab:temperature}, temperature 5 achieves the best or the second-best performance in all shots, so we set the distillation temperature to 5 in all following experiments for COCO.

\underline{Distillation Loss Weight.} With temperature 5, we test the impact of different distillation loss weights. As shown in Table \ref{tab:kd_weight}, the model with weight 5 performs best in all cases, so we select 5 as the weight of distillation loss empirically. Meanwhile, nice results are achieved with weight 10, so we argue that the performance is relatively robust to different weights of distillation loss.

\begin{figure}[t]
    \centering
	{\includegraphics[width=.5\columnwidth]{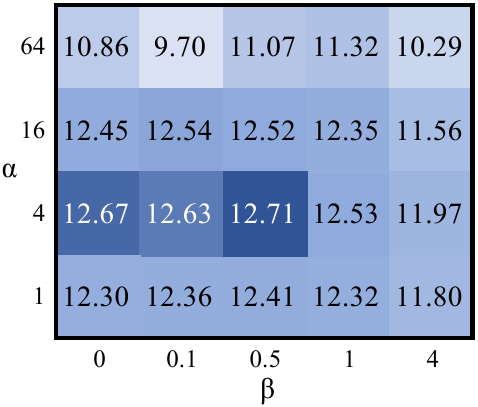}}
	\caption{Experimental results (\%) of the hyper-parameter comparisons of $\alpha$ and $\beta$.}
	\label{fig:weight_heatmap}
\end{figure}

\underline{Weights for FBD.} As shown in Equation \ref{eq:ldr}, there are two parameters to be tuned, $\alpha$ and $\beta$. We carefully explore the impact of their different values and illustrate the results in Figure \ref{fig:weight_heatmap}. We set $\alpha$ to 4 and $\beta$ to 0.5, which achieves the best result.

\begin{table}[t]
	\setlength{\tabcolsep}{5.pt}
	\begin{center}
		\begin{small}
			\begin{tabular}{c|cccccc}
        		\toprule  
        		\multirow{2}*{Temperature} & \multicolumn{6}{c}{Shot Number} \\
        		& 1 & 2 & 3 & 5 & 10 & 30 \\
                \midrule
                1 & 7.00 & 11.10 & 13.18 & 15.32 & 17.85 & 21.59 \\
                \rowcolor{mygray}
                5 & \textbf{7.31} & \textbf{11.56} & \textbf{13.49} & \underline{15.65} & 	\underline{18.13} &	\underline{21.66} \\ 
        		10 & \underline{7.16} & 11.27 &	\underline{13.37} &	\textbf{15.74} & 18.09 & \textbf{21.93}\\
        		20 & 6.95 & \underline{11.43} &	13.22 & 15.45 & 	\textbf{18.27} & \underline{21.66} \\
        		\bottomrule
        	\end{tabular}
		\end{small}
	\end{center}
	\caption{Comparison results (\%) to evaluate the effectiveness of different distillation temperatures on the COCO dataset.}
	\label{tab:temperature}
\end{table}

\begin{table}[t]
	\setlength{\tabcolsep}{5.pt}
	\begin{center}
		\begin{small}
			\begin{tabular}{c|cccccc}
        		\toprule  
        		\multirow{2}*{Weight of KD} & \multicolumn{6}{c}{Shot Number} \\
        		& 1 & 2 & 3 & 5 & 10 & 30 \\
                \midrule
                1 & 7.31 & 11.56 & 13.49 & 15.65 & 	18.13 &	\underline{21.66} \\
                \rowcolor{mygray}
                5 & \textbf{7.69} & \textbf{11.96} & \textbf{13.73} & \textbf{15.82} & \textbf{18.59} & \textbf{21.84} \\ 
        		10 & \underline{7.56} & \underline{11.95} & \underline{13.54} & \underline{15.71} & \underline{18.24} & 21.65\\
        		20 & 7.11 & 11.63 &	13.13 & 14.89 & 18.08 & 21.16 \\
        		\bottomrule
        	\end{tabular}
		\end{small}
	\end{center}
	\caption{FSOD results (\%) of the effectiveness of different distillation loss weights on the COCO dataset.}
	\label{tab:kd_weight}
\end{table}

\begin{table}[t]
	\setlength{\tabcolsep}{4.pt}
	\begin{center}
		\begin{small}
			\begin{tabular}{l|cc|ccccc}
        		\toprule  
        		Teacher & 
        		\multicolumn{2}{c|}{Complexity} & 
        		\multicolumn{5}{c}{Shot Number} \\
        		Model & Time & Space & 1 & 2 & 3 & 5 & 10\\
                \midrule
        		Word2Vec & 0.7s/iter & 9.6G & 56.8 & 63.0 & 64.3 & 64.1 & 66.2 \\
        		GloVe & 0.7s/iter & 9.6G & 56.7 & \underline{63.5} & 64.0 & 63.7 & 65.8  \\
        		CLIP (I+T) & 2.0s/iter & 14.2G & \underline{58.7} & \textbf{64.1} & \textbf{65.6} & \textbf{65.9} & \underline{66.5} \\ \rowcolor{mygray}
                CLIP (T) & 0.7s/iter & 9.6G & \textbf{59.1} & 63.4 & \underline{65.3} & \underline{65.0} & \textbf{66.8} \\ 
        		\bottomrule
        	\end{tabular}
		\end{small}
	\end{center}
	\caption{FSOD results (\%) of the comparisons of our proposed method adopting different teachers on VOC Novel Set 1. The \textit{underlines} denote the second best results. The complexities are measured during training.}
	\label{tab:complex}
\end{table}

\subsection{Discussion on Knowledge Distillation Variants} \label{sec:kdvariant}
As shown in Figure \ref{fig:frame}, D\&R adopts the large-scale pre-trained text encoder as the teacher. The reasons include 1) ``text'' is the semantically-dense data while ``image'' is the semantically-sparse data so that the text encoder teacher can more efficiently improve the student model to explore semantic information from the input data (including images and category texts) than the image encoder teacher, and such efficiency is crucial to the FSOD, which requires to train the model with a few data; 2) the issue of domain shift is far more serious for image data than for text data. Thus, even with the limited data, text encoders can still achieve consistent performance; 3) for vision-language models, the captured knowledge is shared between the text and image encoders, which is due to the training paradigm \cite{DBLP:conf/icml/RadfordKHRGASAM21}. Therefore, the text encoder of vision-language models can teach the student model the general knowledge to capture the semantic knowledge from images; 4) the time and space complexities of adopting the image encoder as the teacher is excessively larger than adopting the text model as the teacher, which demonstrates that it is not worthy of adopting the image encoder as the teacher.

To prove our statements above, we conduct explorations of D\&R by using different teacher variants for knowledge distillation, including CLIP(I+T) having both CLIP image and text encoders as the teachers, CLIP(T) only having the CLIP text encoder, Word2Vec, and GloVe. Note that the last three variants only have the pre-trained text encoder as the teacher. From Table \ref{tab:complex}, we observe that as the comparison between CLIP(T) and CLIP(I+T), the improvement provided by the CLIP image encoder is limited, but the additional time and space consumption is not negligible. Additionally, the main model of D\&R, i.e., CLIP(T), achieves top-2 performance on most tasks. The empirical results demonstrate the proposed statement and D\&R's effectiveness and efficiency.

\section{Conclusion} \label{sec:conclusion}
We introduce the knowledge distillation to FSOD tasks. Then, we discover that the empirical error of the teacher model degenerates the prediction performance of the student model. To tackle this latent flaw, we develop a Structural Causal Model and propose a backdoor adjustment-based knowledge distillation method, D\&R. Empirically, D\&R outperforms state-of-the-art methods on multiple benchmark datasets.

\section*{Acknowledgements}
The authors would like to thank the anonymous reviewers for their valuable comments. This work is supported by the Strategic Priority Research Program of the Chinese Academy of Sciences, Grant No. XDA19020500. The authors are grateful to Hong Wu for the fruitful inspiration.

\bibliography{output}
\end{document}